\documentclass{article}

\usepackage[preprint,nonatbib]{nips_2018}

\usepackage[utf8]{inputenc} 
\usepackage[T1]{fontenc}    
\usepackage{hyperref}       
\usepackage{url}            
\usepackage{booktabs}       
\usepackage{amsfonts}       
\usepackage{nicefrac}       
\usepackage{microtype}      
\usepackage{graphicx}
\usepackage{graphics}
\usepackage[english]{babel}
\usepackage{algorithm}
\usepackage[noend]{algpseudocode}
\usepackage{wrapfig}
\usepackage{amsmath} 
\usepackage{dsfont}
\usepackage[numbers]{natbib}
\usepackage{multirow}
\usepackage{subcaption}
\makeatletter
\algrenewcommand\ALG@beginalgorithmic{\footnotesize}
\makeatother
\title{Gradient Adversarial Training of Neural Networks}

\author{
 Ayan Sinha\thanks{All correspondence to be addressed to asinha@magicleap.com} \\
  Magic Leap\\
  \texttt{asinha@magicleap.com} \\
  \And
  Zhao Chen \\
   Magic Leap\\
  \texttt{zchen@magicleap.com} \\
  \AND
  Vijay Badrinarayanan\\
  Magic Leap\\
  \texttt{vadrinarayanan@magicleap.com} \\
  \And
  Andrew Rabinovich\\
  Magic Leap \\
  \texttt{arabinovich@magicleap.com} \\
}

\begin{document}

\maketitle

\begin{abstract}

We propose gradient adversarial training, an auxiliary deep learning framework applicable to different machine learning problems. In gradient adversarial training, we leverage a prior belief that in many contexts, simultaneous gradient updates should be statistically indistinguishable from each other. We enforce this consistency using an auxiliary network
that classifies the origin of the gradient tensor, and the main network serves as an adversary to the auxiliary network in addition to performing standard task-based training. We demonstrate gradient adversarial training for three different scenarios: (1) as a \emph{defense to adversarial examples} we classify gradient tensors and tune them to be agnostic to the class of their corresponding example, (2) for \emph{knowledge distillation}, we do binary classification of gradient tensors derived from the student or teacher network and tune the student gradient tensor to mimic the teacher's gradient tensor; and (3) for \emph{multi-task learning} we classify the gradient tensors derived from different task loss functions and tune them to be statistically indistinguishable. For each of the three scenarios we show the potential of gradient adversarial training procedure. Specifically, gradient adversarial training increases the robustness of a network to adversarial attacks, is able to better distill the knowledge from a teacher network to a student network compared to soft targets, and boosts multi-task learning by aligning the gradient tensors derived from the task specific loss functions. Overall, our experiments demonstrate that gradient tensors contain latent information about whatever tasks are being trained, and can support diverse machine learning problems when intelligently guided through adversarialization using a auxiliary network.
\end{abstract}

\section{Introduction}
In backpropagation~\cite{backprop} the gradient of the loss function is evaluated with respect to weight tensor in each layer, and and the weights are updated using a learning rule ~\cite{adam}. Gradient tensors recursively evaluated through backpropagation can successfully train deep networks with millions of weight parameters across hundreds of layers and generalize to unseen examples ~\cite{resnet}. However, a mathematical formalism of the generalization ability of deep neural networks (DNNs) trained using backpropagation remains elusive. Indeed, a lack of formalism has given rise to new domains in deep learning such as robustness of DNNs in particular to adversarial examples ~\cite{advexamples}, domain adaptation \cite{DANN}, multi-task learning \cite{relationship}, model compression ~\cite{modelcomp} etc. Here, we investigate the potential of gradient tensors derived during back propagation to serve as an additional cue to learning in these new domains. 

\begin{figure}
\centering
\includegraphics[width=0.88\textwidth]{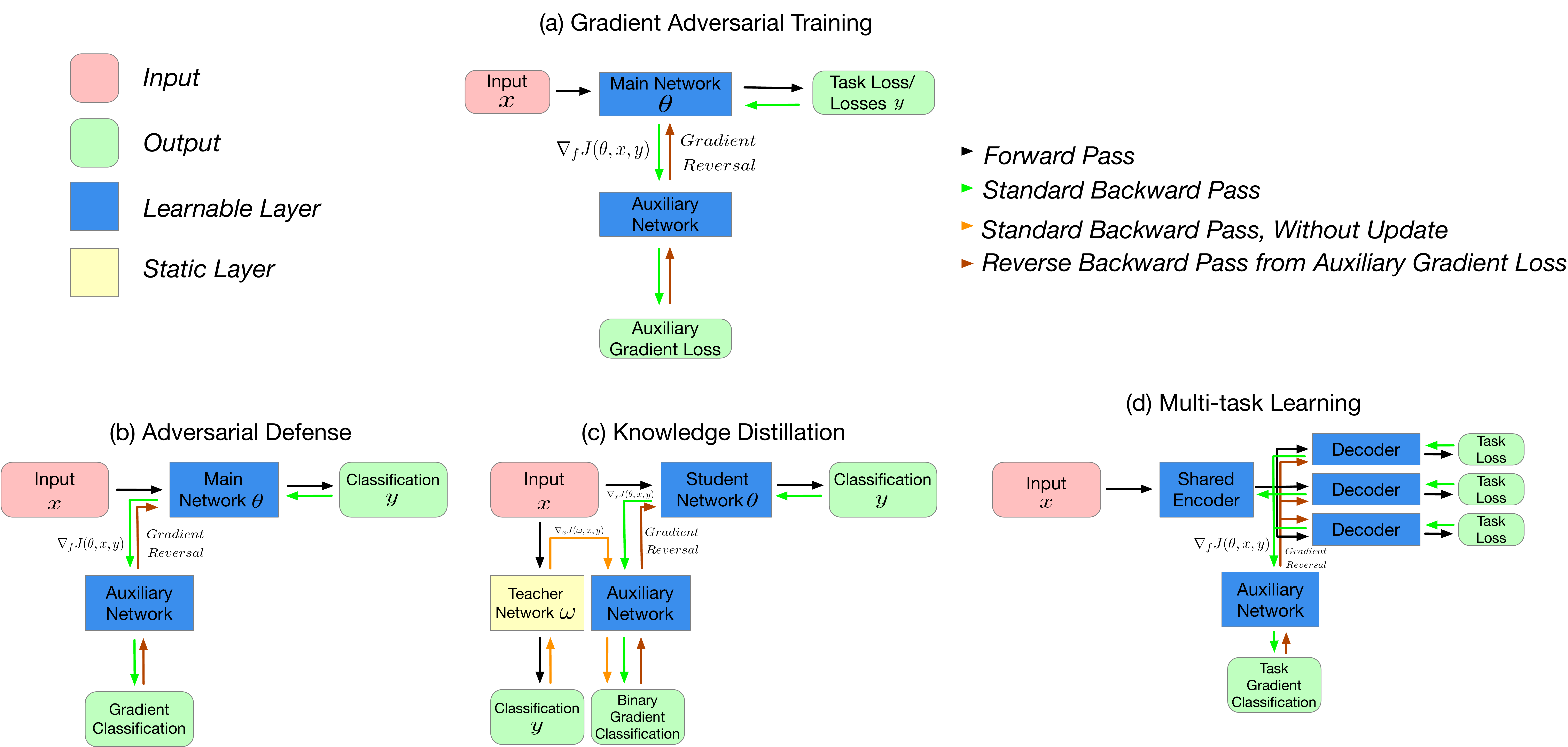}
\caption{\label{fig:mainfig}Gradient Adversarial Training (GREAT) of Neural Networks. The legends on the top left and right of the figure show the information flow in the networks and the different kinds of modules. (a) The general methodology of GREAT wherein the main network is trained using standard backpropagation and also acts as an adversary to the auxiliary network via gradient reversal. The auxiliary network is trained on gradient tensors evaluated during backpropagation. (b) GREAT procedure for adversarial defense: The auxiliary network performs the same classification as the main network, albeit with gradient tensors as input. (c) GREAT method for knowledge distillation: The auxiliary network performs binary classification on the gradient tensors from the student and teacher networks. (d) GREAT method for multi-task learning: The auxiliary networks classifies the gradient tensors from the different task decoders and aligns them through gradient reversal and an explicit gradient alignment layer described later.}
\end{figure}

As demonstrated in prior research, the gradient tensor of the scalar loss function with respect to the input or intermediate layer, termed  the Jacobian $J$,  is highly informative~\cite{gradcam}. This follows naturally from the equations of backpropagation for a perceptron,
\begin{equation}
 \delta^l=(w^{l+1})^T\delta^{l+1}\odot \sigma'(z^l),
\end{equation}
with $\delta^L=\nabla_a C \odot \sigma'(z^L)$.  Here $\delta$ is the gradient tensor, $l$ is the layer with $L$ being the final layer, $\nabla_a C$ is the gradient of loss function with respect to the neural network output $a$ after the final activation, $\sigma$ is the activation function, $z^l$ is the output after layer $l$ with $a=\sigma'(z^L)$, $w$ is the weight matrix, and $\odot$ is the Hadamard product. It follows from these equations that the gradient tensor at any layer is a function of both the loss function and all \emph{succeeding} weight matrices. The information from gradient tensors have been employed classically for regularization~\cite{doubleback} and more recently for visualizing saliency maps~\cite{salmaps}, interpreting DNNs~\cite{visualizing,allconvnet}, generating adversarial examples~\cite{harnessing} and weakly supervised object localization\cite{gradcam}. Most approaches use the information from the gradient tensor in a separate step to achieve the desired quantitative or qualitative result. Different from these approaches, we use the gradient tensor during the training procedure via an adversarial process \cite{GAN} in our proposed GRadiEnt Adversarial Training (GREAT) procedure.

The main premise underlying GREAT is that the information in the gradient tensor inhibits reliable training dynamics under certain scenarios. GREAT aims to nullify the \emph{dark information} in the gradient tensors by first processing the gradient tensor in an auxiliary network and then passing an adversarial signal back to the main network (Figure \ref{fig:mainfig}a) via the gradient reversal procedure~\cite{DANN}. This adversarial signal regularizes the weight tensors in the main network akin to double backpropagation~\cite{doubleback}. Using calculus, the adversarial gradient signal $\varrho$ flowing forward in the main network can be shown to be,
\begin{equation}
 \varrho^{l+1}=-w^{l+1}\varrho^{l}\odot \sigma'(z^l),
\end{equation}
which is of a similar functional form as $\delta$ but of opposite sign and affected by \emph{preceding} weight matrices till the layer of the considered gradient tensor. As networks tend to have perfect sample expressiveness as soon as the number of parameters exceeds the number of data points~\cite{rethinkgen}, we expect the regularization provided by the auxiliary network to improve robustness and not considerably affect performance. We describe the \emph{dark information} present in the gradient tensors in three scenarios: (a) adversarial examples, (b) multi-task learning, and (c) knowledge distillation~\cite{distillation}. We describe the intuition behind using GREAT for these three scenarios in the subsequent paragraphs and describe the exact training methodology in Section 3. 

Adversarial examples are carefully crafted perturbations applied to normal images which are usually imperceptible to humans, but can seriously confuse state-of-the-art deep learning models~\cite{advexamples,harnessing}. A common step to all adversarial example generation is calculating the gradient of the objective function with respect to the input~\cite{madry} called the saliency map. The objective function is either the task loss function or derived  from it. This gradient tensor is processed to perturb the original image, and the model mis-classifies the perturbed image. We use GREAT to make the saliency maps uninformative (Figure \ref{fig:mainfig}b), and hence, mitigate the network's susceptibility to adversarial examples. 

The objective of knowledge distillation is to compress the predictive behavior of a cumbersome DNN (teacher) or an ensemble of DNNs into a simpler model (student)~\cite{distillation,ensembling}. Distilling knowledge to a student network is achieved by matching the logits or soft output distribution of the teacher to the output of the student in addition to usual supervised loss function. In Figure \ref{fig:mainfig}c, we show how GREAT provides a complementary approach to distillation wherein we statistically match the gradient tensor of the teacher to the student using the auxiliary network, in lieu of matching output distributions.

In multi-task learning, a single network is trained end-to-end to achieve multiple related but different task outputs for an input~\cite{relationship}. This is achieved by having a common encoder and separate task-specific decoder. In a perfect multi-task learning scenario, the gradient tensors of the individual task-loss functions with respect the the last shared layer in the encoder should be indistinguishable so as to coherently train all the shared layers in the encoder. We use GREAT to train a \emph{gradient alignment layer} between the encoder and task-specific decoders which operates in the backward pass so that the task-specific gradient tensors are less distinguishable by the auxiliary network (Figure \ref{fig:mainfig}d). 

In Section 2, we describe the GREAT procedure for each of the above scenarios. In Section 3, we highlight the results of GREAT and in Section 4 we discuss conclusions and possible avenues of future work. Note we discuss relevant work as appropriate in the remainder of this article. 

 \begin{wrapfigure}{r}{0.41\textwidth}
  \begin{center}
    \includegraphics[width=0.40\textwidth]{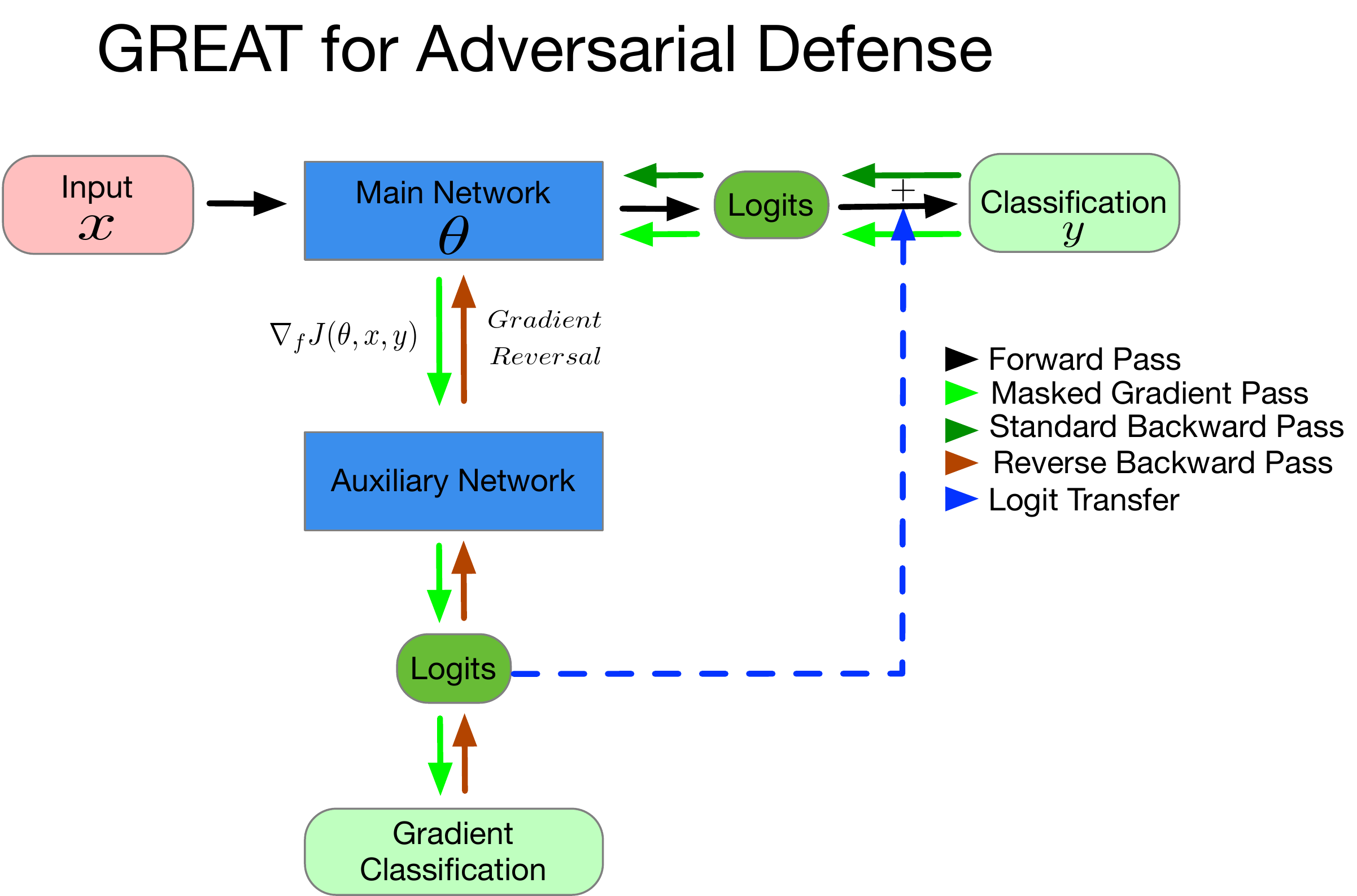}
  \end{center}
  \caption{\label{advdeffig}Adversarial defense comprised of GREAT and GREACE. In GREACE, the output probability distribution from auxiliary network is added to the gradient of the loss with respect to the logits in the main network to help separate negative classes whose gradient tensors are similar to primary class.}
\end{wrapfigure}
\section{Gradient Adversarial Training}
We describe the adaptations of GREAT suitable for adversarial defense, knowledge distillation and multi-task learning.  
\subsection{Adversarial defense}
The general objective for defense against adversarial examples is
\begin{equation}\label{eqadv}
 J(\theta,x,y)=J(\theta,x+\Delta,y) \quad, \quad  \forall \quad\|\Delta\|_p\leq \epsilon .  
\end{equation}
Here, $x$ is the input, $y$ the output, $\theta$ the network parameters, $\Delta$ is the perturbation tensor whose $p$-norm is constrained to be less than $\epsilon$, and $J$ subsumes the loss function and the network architecture. Non-targeted attacks are devised by $x +\epsilon f(\nabla J(\theta,x,y))$, i.e., moving in the direction of the gradient of the ground truth class $y$, where $f$ is usually the sign function in FGSM; whereas targeted attacks are calculated as $x -\epsilon f(\nabla J(\theta,x,\bar y))$ for $\bar y \neq y$. Using first order Taylor series approximation in equation \ref{eqadv} amounts to the equivalent formulation,
\begin{equation}\label{eqadveq}
 \nabla J(\theta,x+\Delta,y)\Delta\approx0 \quad, \quad  \forall \quad\|\Delta\|_p\leq \epsilon.  
\end{equation}
Previous attempts at adversarial defenses have focused on minimizing $\|\nabla J(\theta,x+\Delta,y)\|_p$ locally at the training points~\cite{defensivedistill,Ross2017ImprovingTA, datagrad,advrob}. However, this leads to a sharp curvature of the loss surface near those points, violating the first order Taylor approximation, which in turn makes the defense ineffective \cite{certified}. 

\textbf{GREAT:} Our GREAT procedure removes the class-specific information present in the gradient tensor. Formally, for all $N$ samples in the training set,
\begin{equation}\label{greatadv}
 \nabla J(\theta,x_i,y_i)= \nabla J(\theta,x_i,\hat y_i) \quad, \quad \forall \\\hat y_i\neq y_i\quad \textrm{and}\quad \forall \\ i\in N.  
\end{equation}
In the absence of class-specific information, a single-step targeted attack becomes hard as the perturbation tensor is class-agnostic. However, GREAT makes the gradient tensors class-agnostic or in other words obfuscates the gradient. Networks with obfuscated gradients are still vulnerable to sophisticated iterative attacks \cite{obgrad} and to universal adversarial perturbations \cite{universalap}. Hence, as a second line of defense we propose gradient-adversarial cross-entropy (GREACE) loss. 

\textbf{GREACE:} GREACE adapts the cross-entropy loss function to add weight to the negative classes whose gradient tensors are similar to those of the primary class. The weight is added to the negative classes in the gradient tensor flowing backward from the soft-max activation, before back-propagating through the rest of the main network (see Figure \ref{advdeffig}). The weight is evaluated using the soft-max distribution from the auxiliary network which indicates the similarity of gradient tensor of the primary class to the negative classes. This added weight helps separate the high-dimensional decision boundary between easily confused classes, similar in spirit to confidence penalty~\cite{confidencepen} and focal loss~\cite{focalloss}, albeit from the perspective of gradients. Mathematically, the gradient tensor from the cross-entropy loss is modified in the following way, 
\begin{equation}\label{advce}
 \nabla_a \hat C \mapsto \nabla_a C + \beta*\sigma(\acute{a})\mathds{1}_{\hat y\neq y}. 
\end{equation}
Here, $\hat C$ and $C$ are the GREACE and original cross-entropy functions respectively, $a$ and $\acute{a}$ are the output activations from the main and auxiliary network respectively, $\sigma$ is the soft-max function, $\beta$ is a penalty parameter, and $\mathds{1}_{\hat y\neq y}$ is a one-hot function for all $\hat y$ not equal to the original class $y$, i.e., negative classes. The gradient fed into the auxiliary network is masked after passing through the soft-max function in the main network,  $\nabla C_a\mathds{1}_{y}$. This avoids the auxiliary classifier to catch onto gradient cues from negative classes and only concentrates on the class in question. We also experimented with the unmasked gradient tensor, but the results weren't as good. The combined objective for adversarial defense is:
\begin{equation}\label{objdef}
\textrm{min}_{\theta} \hat{J}(\theta,x,y) + \alpha~\textrm{max}_{\acute\theta}  {J}(\acute{\theta} ,\nabla \bar J(\theta,x,y),y).
\end{equation}
$\hat{J}$ indicates the GREACE, $J$ indicates the standard cross-entropy, $\bar J$ indicates the masked cross-entropy, and $\alpha$ is a weight parameter for the auxiliary network's loss.  
\begin{algorithm}[t]
\caption{Algorithm for defense against adversarial examples using GREAT and GREACE}\label{algadv}
\begin{algorithmic}[1]
\Procedure{Train}{$\theta,\acute{\theta}$}\Comment{Requires inputs $x$, labels $y$, penalty $\beta$}
\While{$j<j_{max}$}\Comment{$j$ is current iteration}
\State $C\gets {J}(\theta ,x,y)$ \Comment{Main network loss by forward pass}
\State $g \gets \nabla \bar J(\theta,x,y)$ \Comment{Evaluate masked gradient tensor w.r.t. $x$}
\State $\acute C\gets {J}(\acute{\theta} ,g,y)$ \Comment{Auxiliary network loss by forward pass}
\State $\acute{\theta}(j) \to \acute{\theta}(j+1) $ \Comment{Update weights in auxiliary network using $\acute C$}
\State $\nabla \acute C_g \gets -\alpha\nabla {J}(\acute{\theta} ,g,y) $\Comment{Evaluate reversed gradient w.r.t $g$}
\State $ \nabla_a \hat C \gets \nabla_a C + \beta*S(\acute{a})\mathds{1}_{\hat y\neq y}$ \Comment{Evaluate GREACE loss}
\State ${\theta}(j) \to {\theta}(j+1) $ \Comment{Update weights in main network using $\nabla_a \hat C,\nabla_g \acute C$}

\EndWhile\label{euclidendwhile}
\EndProcedure
\end{algorithmic}
\end{algorithm}

\subsection{Knowledge distillation}
In classical distillation~\cite{distillation} the student's output distribution $S(x)$ mimics the teacher's soft output distribution $T(x)$. In GREAT, the student model mimics teacher model's gradient distribution which is a weaker constraint as it allows final distributions to differ by a constant value. A solution for student, $S(x)$ which jointly minimizes the supervised loss and $\nabla S(x) = \nabla T(x)$ exists, as proved in \cite{sobolev}. GREAT uses a discriminator to match the gradient distributions owing to the success of adversarial losses \cite{GAN} over traditional regression-based loses.
\begin{algorithm}[t]
\caption{Algorithm for knowledge distillation using GREAT procedure}\label{algkd}
\begin{algorithmic}[1]
\Procedure{Train}{$\theta,\acute{\theta}$}\Comment{Requires inputs $x$, labels $y$, teacher $T$ with parameters $\tau$}
\While{$j<j_{max}$}\Comment{$j$ is current iteration}
\State $C\gets {J}(\theta ,x,y)$ \Comment{Student network loss by forward pass}
\State $g_s \gets \nabla J(\theta,x,y)$ \Comment{Evaluate student gradient tensor w.r.t. $x$}
\State $g_t \gets \nabla J(\tau,x,y)$ \Comment{Evaluate teacher gradient tensor w.r.t. $x$}
\State $\acute C\gets \hat{J}(\acute{\theta} ,g_t,1)+\hat{J}(\acute{\theta} ,g_s,0)$ \Comment{Binary classifier loss by forward pass}
\State $\acute{\theta}(j) \to \acute{\theta}(j+1) $ \Comment{Update weights in auxiliary network using $\acute C$}
\State $\nabla C \gets (1-\alpha)\nabla J(\theta,x,y)$ \Comment{Evaluate gradient tensor of loss}
\State $\nabla_{g_s} \acute C \gets -\alpha\nabla {\hat J}(\acute{\theta} ,g_s,0) $\Comment{Evaluate reversed gradient w.r.t $g_s$}
\State ${\theta}(j) \to {\theta}(j+1) $ \Comment{Update weights in main network using $\nabla C, \nabla_{g_s} \acute C$}
\EndWhile
\EndProcedure
\end{algorithmic}
\end{algorithm}
The GREAT procedure for knowledge distillation mimics a GAN training procedure. The binary classifier discriminates between student and teacher model gradients and drives the student model to generate gradient tensor distribution similar to the teacher model as shown in Figure \ref{fig:mainfig}c. The objective to be optimized is:
\begin{subequations}\label{eq1}
\begin{eqnarray}
&(1-\alpha)~\textrm{min}_{\theta} J(\theta,x,y) + \alpha~\textrm{min}_{\theta} \textrm{max}_{\omega} D(\theta, \omega, x,y)\\
&D(\theta, \omega, x,y) = E_{t\sim \nabla T(x)}\textrm{log} f(t,\omega) + E_{s\sim \nabla  J(\theta,x,y)}\textrm{log}(1-f(s,\omega)).
\end{eqnarray}
\end{subequations}
$f$ is the binary classifier with $\omega$ parameters, $s,t$ are gradient tensors from the student and teacher, respectively, $E$ denotes expectation, and $\alpha$ is a loss balancing parameter. GREAT has no hyper-parameter controlling the teacher's distribution to be matched, unlike the hard to set temperature parameter in ~\cite{distillation}. However, we have an extra pass through the student network. 

\subsection{Multi-task learning}
\begin{wrapfigure}{r}{0.43\textwidth}
  \begin{center}
    \includegraphics[width=0.42\textwidth]{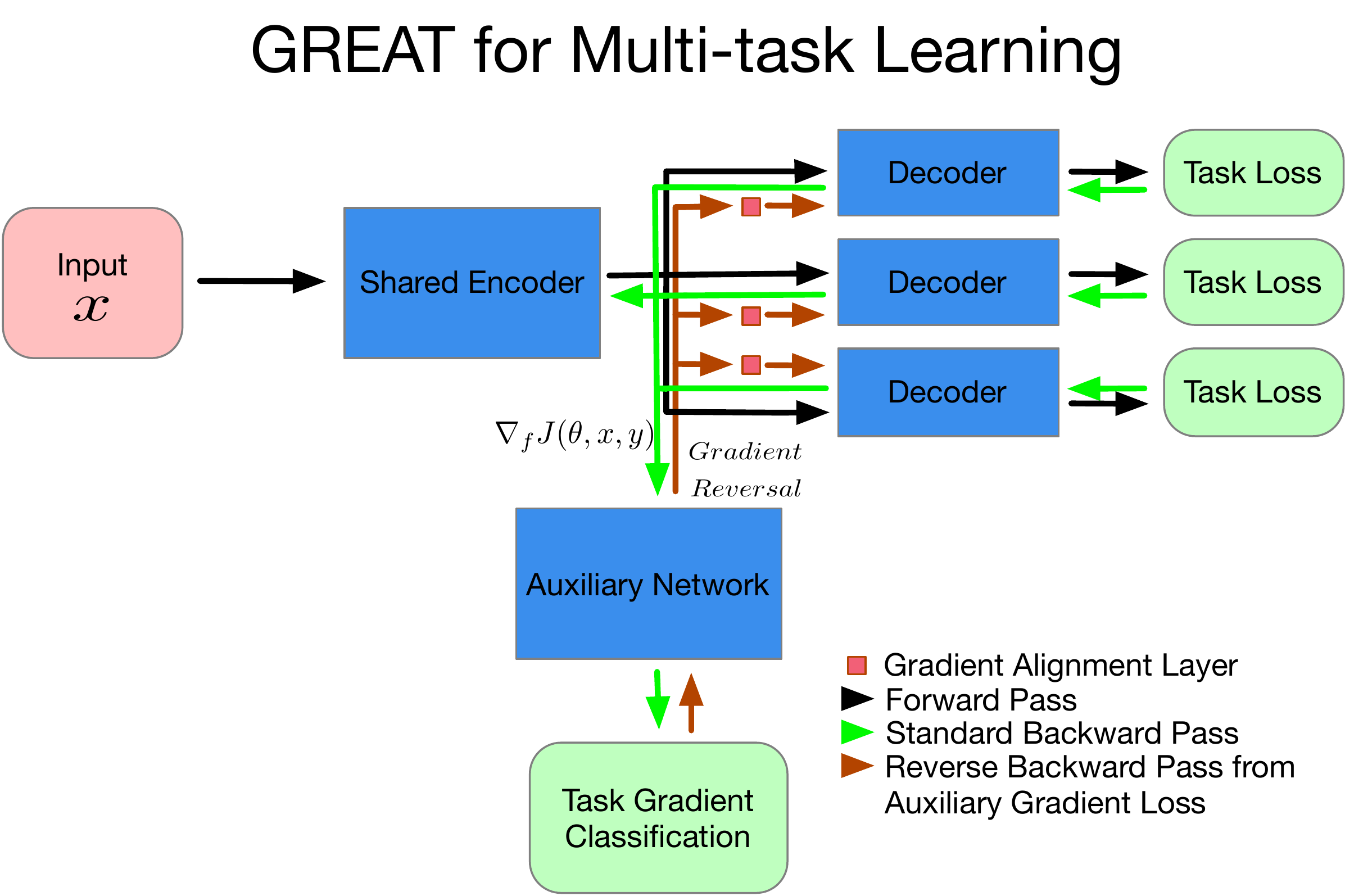}
  \end{center}
  \caption{\label{mlt} GREAT for multi-task learning. The GALs are trained using reversed gradients from the auxiliary task classifier.}
\end{wrapfigure}

GradNorm ~\cite{gradnorm} adaptively balances the loss-weights based on the norm of the gradients. The GREAT procedure for multi-task learning can be viewed as a generalization of GradNorm with two important differences: (1) We do not enforce that the gradients have balanced norms, but instead, desire that they have similar statistical distributions. This is achieved by the auxiliary network similar to a discriminator in a GAN setting. (2) Instead of assigning task-weights, we add extra-capacity to the network in the form of gradient-alignment layers (GALs). These layers are placed after the shared encoder and before each of the task-specific decoders as shown in Figure \ref{mlt}. They have the same dimensions as the last shared feature tensor minus the batch size, and are active only during the backward pass, i.e., the GALs are dropped during forward inference. 
\begin{algorithm}
\caption{Algorithm for multi-task learning using GREAT on GALs}\label{algml}
\begin{algorithmic}[1]
\Procedure{Train}{$\theta,\acute{\theta},\omega_i,\gamma_i $}\Comment{Requires inputs $x$, labels for tasks $y_i$}
\State $\gamma_i\gets \mathds{1},\quad C_i^0\gets {J_i}(\theta ,\omega_i,x,y_i)$ \Comment{Initialize GAL tensors with ones and initial task losses}
\While{$j<j_{max}$}\Comment{$j$ is current iteration}
\State $C_i\gets {J_i}(\theta ,\omega_i,x,y_i)/{C_i^0}\quad \forall i $ \Comment{Normalize task losses after forward pass}
\State $g^f_i \gets \nabla J_i(\omega_i,x,y_i) \quad \forall i $ \Comment{Evaluate task gradient tensors w.r.t. feature $f$}
\State ${\omega_i}(j) \to {\omega_i}(j+1) \quad \forall i $ \Comment{Update weights in decoders using $\nabla C_{i}$}
\State ${\theta}(j) \to {\theta}(j+1) $ \Comment{Update weights in encoder using $\sum_i g^f_i \gamma_i$ }
\State $\acute C\gets \acute{J}(\acute{\theta} ,g^f_i\gamma_i,\acute y)$ \Comment{Task classification loss by forward pass}
\State $\acute{\theta}(j) \to \acute{\theta}(j+1) $ \Comment{Update weights in task classifier network using $\acute C$}
\State $\nabla_{\gamma_i} \acute C \gets -\nabla\acute{J}(\acute{\theta} ,g^f_i\gamma_i,\acute y) $\Comment{Evaluate reversed gradient w.r.t $\gamma_i$}
\State ${\gamma_i}(j) \to {\gamma_i}(j+1) \quad \forall i $ \Comment{Update weights in GALs using $\nabla_{\gamma_i}\acute C$}
\EndWhile
\EndProcedure
\end{algorithmic}
\end{algorithm}

The auxiliary network receives the gradient tensor from each task as input and classifies them according to task. Successful classification implies the gradient tensors are discriminative, which impedes training of the shared encoder as the gradients are misaligned. The GALs mitigate the misalignment by element-wise scaling of the gradient tensors from all tasks. These layers are trained using the reversed gradient signal from the auxiliary network, i.e., the GALs attempt to make the gradient tensors indistinguishable. Intuitively, the GALs observe the statistical irregularities that prompt the auxiliary classifier to successfully discriminate between the gradient tensors, and then adapt the tensors to remove the irregularities or equalize the distributions. Note, that the task losses are normalized by the initial loss so that the alignment layers are tasked with local alignment and not global loss scale alignment. Furthermore, the soft-max activation function in the auxiliary network's classification layer implicitly normalizes the gradients. The values in the GAL weight tensors are initialized with ones and restricted to be positive for training to converge. In practice, we observed that a low learning rate ensured positivity of the GAL tensors. The overall objective for multi-task learning is:

\begin{equation}\label{objml}
\textrm{min}_{\theta,\omega_{1, \cdots N}}\textstyle\sum_i J_i(\theta,\omega_i,\gamma_i, x,y_i) + \textrm{max}_{\acute\theta,\gamma_{i\cdots N}}  {\acute J}(\acute{\theta} ,\nabla J_i(\omega_i,x,y_i)\gamma_i,\acute y)
\end{equation}
$J_i$ are normalized task losses, $ \acute J$ is N-class cross-entropy loss, $\theta, \acute{\theta}$ are learnable parameters in  shared encoder and auxiliary classifier, respectively,  $\omega_i, \gamma_i, y_i$ are decoder parameters, GAL parameters, and labels for task $i$ respectively, and $\acute y$ represent the task labels.  
\begin{table}[t]
  \centering
  \renewcommand{\arraystretch}{1}
  \scalebox{0.95}{
  \begin{tabular}{|c|c|c|c|c|c|c|c|c|}
    \hline
    \multirow{3}{1cm}{\textbf{Method}} & \textbf{Train}  & \textbf{No-Attack} & \multicolumn{2}{c|}{\textbf{Non-Targeted}} & \multicolumn{4}{c|}{\textbf{Targeted}}\\
    \cline{4-9}
    & & &\textbf{FGSM} & \textbf{iFGSM} & \multicolumn{2}{c|}{\textbf{FGSM}} &\multicolumn{2}{c|} {\textbf{iFGSM}}\\
    \cline{6-9}
    & & & & & Worst & Random& Worst& Random\\  
    \hline
Baseline  &	99.97 &	\textbf{93.32} &	32.75&1.99 & 72.89&10.43 & 89.59&  18.29\\
Adversarial &	99.97 &	89.91	& 56.88 & 16.73 & 82.07 & 45.26 & \textbf{89.81} & 69.89\\
GREACE &	92.56 &	89.84	&77.90 & 72.40 & 83.39 & 66.23 & 87.13 & 79.10  \\
GREAT &	99.53 &	91.95	&47.51 & 15.45 & 72.73 & 12.78 & 89.62 &  21.95 \\
GRE(AT+CE) &	90.87 & 89.97 & \textbf{81.28} & \textbf{77.04} & \textbf{84.53} & \textbf{73.52} & 88.57 & \textbf{82.38}\\
\hline
  \end{tabular}}
    \vspace{1pt}
  \caption{\label{cifar}CIFAR-10 test accuracy in \% values of different training methods on targeted and non-targeted attacks using FGSM and iFGSM. We use $\epsilon=0.1$, and set $k=10$ for iFGSM. GREA(AT+CE) is the best defense for all but one adversary highlighting the importance of gradient adversarial training in addition to GREACE during training.}
  
\end{table}
\begin{table}[t]
  \centering
  \renewcommand{\arraystretch}{1}
    \scalebox{0.95}{
  \begin{tabular}{|c|c|c|c|c|c|c|c|c|}
    \hline
    \multirow{3}{1cm}{\textbf{Method}} & \textbf{Train}  & \textbf{No-Attack} & \multicolumn{2}{c|}{\textbf{Non-Targeted}} & \multicolumn{4}{c|}{\textbf{Targeted}}\\
    \cline{4-9}
    & & &\textbf{FGSM} & \textbf{iFGSM} & \multicolumn{2}{c|}{\textbf{FGSM}} &\multicolumn{2}{c|} {\textbf{iFGSM}}\\
    \cline{6-9}
    & & & & &Worst & Random& Worst& Random\\
    \hline
Baseline  &	99.97 &	\textbf{96.20} &	45.97 & 6.26 & \textbf{96.20} & \textbf{96.20} & 80.27 & 15.65\\
Adversarial & 99.98 &94.92	 &	58.70 & 8.41 & 94.92 & 94.92 & 83.70 & 19.19\\
GREACE &	93.52 &	93.71	&	73.65 & \textbf{80.13} & 93.70 & 93.70 & 89.26 & \textbf{88.09}\\
GREAT &	99.76 &	95.58	& 45.95 & 5.95 & 95.58 & 95.58 & 79.09 &15.68 \\
GRE(AT+CE) & 92.95   &	93.90 & \textbf{74.12} & 79.36 & 93.89 & 93.90 & \textbf{89.56} & 87.37\\
\hline
  \end{tabular}}
  \vspace{1pt}
  \caption{\label{svhn}SVHN test accuracy in \% values of different training methods on targeted and non-targeted attacks using FGSM and iFGSM. We use $\epsilon=0.2$, and set $k=10$ for iFGSM. GREA(AT+CE) or GREACE is the best defense for all iFGSM and non-targeted FGSM adversaries showing that the modified cross entropy loss robustly separates classes.}
\end{table}

\section{Results}
\subsection{Adversarial defense}
\begin{figure}
  \centering
      \begin{subfigure}{0.65\textwidth}
	\includegraphics[width=0.99\textwidth]{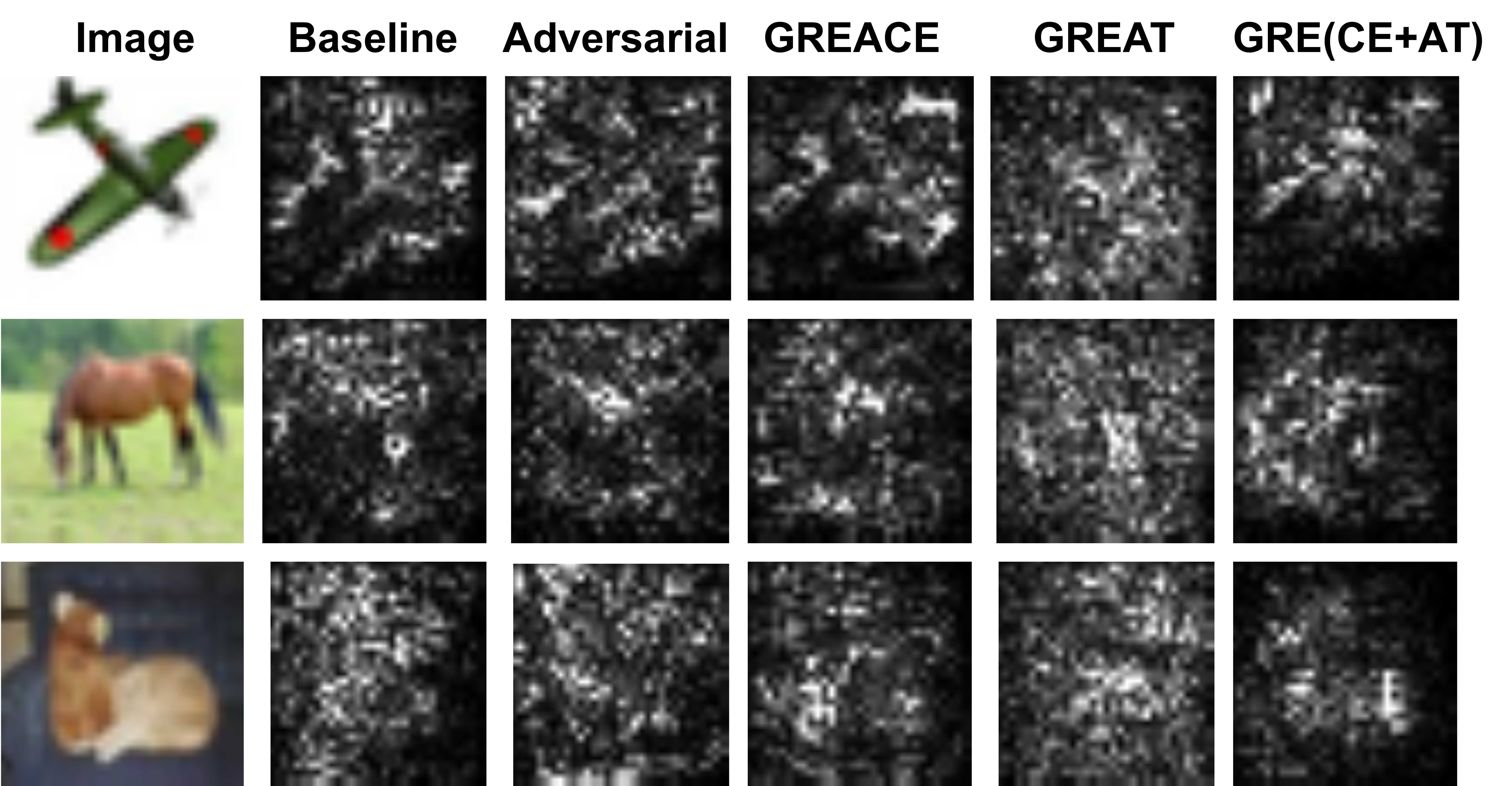}
    \end{subfigure}
  \begin{subfigure}{0.25\textwidth}
	\includegraphics[width=0.99\textwidth]{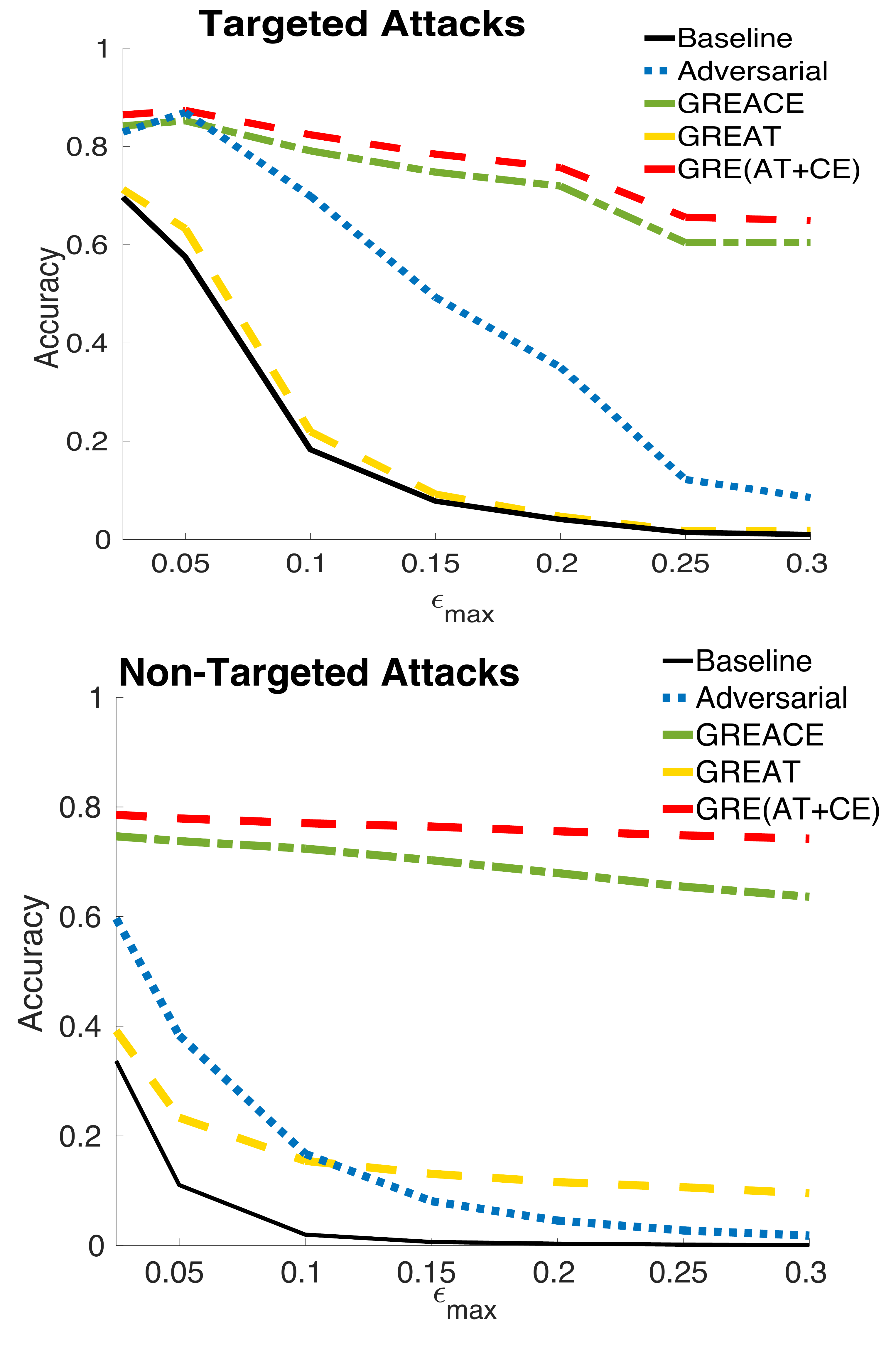}
    \end{subfigure}
  \caption{\label{figsal} \emph{Left}: Saliency maps for images in CIFAR-10 for different training methods. Baseline and adversarial are active outside object, GREACE is sparse, GREAT is uninformative, GRE(AT+CE) is sparse, less informative and active within object.  \emph{Right}: Accuracy plots of different methods against adversaries with maximum allowed perturbation, $\epsilon$. GREAT (yellow) is more robust than Adversarial (blue) to non-targeted adversaries for high $\epsilon$. GREACE (green) and GRE(AT+CE) (red) are uniformly robust for different $\epsilon$.}
\end{figure}

We demonstrate GREAT on the CIFAR-10 and SVHN datasets. We use a ResNet-18 architecture \cite{resnet} for both datasets. We observed that ResNet models are more effective in the GREAT training paradigm for adversarial defense relative to models without skip connections. In GREAT, skip connections help propagate the gradient information in the usual backward pass, as well as forward propagate the reversed gradient from the auxiliary classifier network through the main network. In our experiments, the auxiliary network is a copy of the main network. We gradually increase the auxiliary loss weight parameter, $\alpha$ and the penalty parameter, $\beta$ to their final values, $\alpha_{max},\beta_{max}$ so as to not impede the main training task during initial epochs. We empirically set $\alpha_{max}=1$  and $\beta_{max}$ to 2 and 10 for CIFAR-10 and SVHN, respectively. These values optimally defend against adversarial examples, while not adversely affecting the test accuracy on the original samples. The network architectures and additional parameters are discussed in the supplement. We evaluate our method against targeted and non-targeted adversarial examples using the fast gradient sign method (FGSM) and its iterated version (iFGSM) for $k$ iterations. For targeted attacks we report the test accuracy for adversaries choosing a random target class or the worst (least probability) target class. We compare our method against adversarial training and base network with no defense mechanism in Tables~\ref{cifar} and~\ref{svhn}. We employ FGSM adversaries in the adversarially trained network, described further in the supplement. Most other defenses are not effective as reported in~\cite{obgrad}. For CIFAR-10, we also draw plots of the test accuracy as a function of the maximum perturbation, $\epsilon$ allowed by the adversary in Figure~\ref{figsal}. Firstly, the training set accuracy indicates that GREACE acts as a strong regularizer, and the combination of GREACE and GREAT prevents over-fitting to the training set. Second, we see that GREAT adds robustness to non-targeted single-step attacks but fails against iterated adversary (iFGSM), an indication of gradient obfuscation~\cite{obgrad}. Third, we see that GREACE in isolation is robust to adversarial attacks, however, the combination of GREAT and GREACE boosts robustness. Surprisingly, GRE(AT+CE) performs better than adversarial training on single step attacks, even though adversarial training is trained to be robust against them. Finally, in Figure~\ref{figsal} we see that the performance of GRE(AT+CE) deteriorates slightly for strong adversaries with high $\epsilon$ values validating the robustness of the classifier. \footnote{Single-step targeted attacks are not successful on SVHN due to the simple task of recognizing digits}. The saliency maps for the different methods are plotted in Figure~\ref{figsal} for 3 examples of CIFAR-10. Pixel activations around an object promote generation of adversarial examples. We see that the saliency maps for baseline and adversarial training have high pixel activations both within and around the object, whereas activations for GREAT are very noisy and not discriminative as expected. In contrast, the saliency maps for GRE(CE+AT) are sparse and predominantly activated within the object, hence, mitigating adversarial examples.   

\begin{table}[b]
  \centering
  \begin{tabular}{|c|c|c|c|c|c|c|c|c|c|c|}
    \hline
    \multirow{3}{1cm}{\textbf{Method}}  & \multicolumn{4}{c|}{\textbf{CIFAR-10}} & \multicolumn{4}{c|}{\textbf{mini-ImageNet}}\\
    \cline{2-9}
     &\multicolumn{2}{c|}{\textbf{CNN(S)+RN(T)}} & \multicolumn{2}{c|}{\textbf{RN(S)+RNx(T)}} &\multicolumn{2}{c|} {\textbf{RN(S)+RN152(T)}}&\multicolumn{2}{c|} {\textbf{RN(S)+RN152(T)}}\\
    \cline{2-9}
     &100\% &5\%& 100\% & 5\% & 100\%&5\%& 100\%&5\%\\
    \hline
Baseline  &	84.74&	65.41 &	93.19	& 66.73&59.24 & 14.41&\textbf{58.02}&13.79\\
Distillation &	85.69&	66.45 &	\textbf{93.65}	& 67.69&51.72 & 16.73&46.77&14.00\\
GREAT &	\textbf{85.72}&	\textbf{66.55} &	93.43	& \textbf{67.80}&\textbf{59.80} &\textbf{16.82}&56.31&\textbf{14.02}\\
\hline
  \end{tabular}
  \caption{\label{kd}Results of knowledge distillation on CIFAR-10 and mini-ImageNet. RN refers to ResNet-18. The third row indicates the \% of all train samples used during training. GREAT performs best in the sparse regime for all combinations and better than distillation on all but 1 scenario. }
\end{table}

\subsection{Knowledge distillation}
We demonstrate GREAT's potential for knowledge distillation on the CIFAR-10 and mini-ImageNet datasets. The mini-ImageNet dataset is a subset of the original ImageNet dataset with 200 classes, and 500 training and 50 test samples for each class. We show distillation results for 2 scenarios: (a) all training examples are used to train the student model, i.e, dense regime and (b) only 5\% of training samples are used to train the student models, i.e., sparse regime. For CIFAR-10, we use (i) a 5-layer CNN and a pretrained ResNet-18, and (ii) ResNet-18 and a pretrained ResNext-29-8\cite{resnext} as student-teacher combinations. For mini-ImageNet, we train a teacher ResNet-152 model at two resolutions: (i) 64x64, (ii) 224x224 for 100 epochs and 50 epochs, respectively. We use a ResNet-18 as the student model at both resolutions. We use a shallower version of the student model as the auxiliary binary classifier. Details of the architecture, optimizer and learning rate policy for each scenario are in the supplement. We compare GREAT against a baseline model trained using cross-entropy loss, and against a distilled model trained using a combination of cross-entropy and unsupervised KL-loss. We determined the best temperature and $\alpha$ parameter for distillation in the two training regimes on the 5-layer CNN+ResNet-18 combination, and used these parameters for the mini-ImageNet experiments. The optimal parameters were chosen through grid search for the ResNet-18+ResNext-29-8 combination. We set $\alpha=0.1$ in all experiments using GREAT determined from the dense training regime of CNN+ResNet-18 combination. The results are reported in Table~\ref{kd}. We see that GREAT consistently performs better than the baseline and distillation in the sparse training regime, indicating better regularization by the gradient adversarial signal. The baseline model performs best for the full resolution, dense training regime for mini-ImageNet indicating that the teacher model trained for only 50 epochs provides weak learning cues. Indeed, the best test accuracy reported for mini-ImageNet at full resolution is 83.32\% as opposed to our teacher model with 71.30\% top-1 accuracy. The poor performance of distillation on mini-ImageNet dense regime indicate that the hyper parameters determined on CIFAR-10 are not transferable across datasets. In contrast, GREAT with the same $\alpha$ parameter is able to coherently distill the model for both the dense and sparse training regimes across different student-teacher combinations. 

\begin{table}[b]
  \centering
  \begin{tabular}{|c|c|c|c|c|c|c|c|c|c|}
    \hline
    \multirow{3}{1cm}{\textbf{Method}}  & \multicolumn{4}{c|}{\textbf{CIFAR-10}} & \multicolumn{3}{c|}{\textbf{NYUv2}}\\
    \cline{2-8}
     &\textbf{Class}  &\textbf{Color} & \textbf{Edge} & \textbf{Auto}& \textbf{Depth}& \textbf{Normal}& \textbf{Keypoint}\\
    \cline{2-8}
     &\% Error  &RMSE& RMSE & RMSE& RMSE& 1-|cos| &RMSE\\
    \hline
Equal  &	24.0 &		0.131 &	0.349 & 0.113&0.861 &		0.207	&	0.407 \\
Uncertainty &	26.6 &		\textbf{0.111}	&	0.270 &0.090& 0.796 &		0.192	&	0.389 \\
GradNorm &	\textbf{23.5} &		0.116	&	0.270 & 0.091&0.810 &		0.169	&	\textbf{0.377} \\
GREAT &	24.2 &		0.114	&	\textbf{0.252} & \textbf{0.087} &\textbf{0.779} &		\textbf{0.167}	&	0.382 \\
\hline
  \end{tabular}
  \caption{\label{multitask} Test errors of multi-task learning on the CIFAR-10 and NYUv2 datasets. GREAT performs best on 2 tasks each for CIFAR and NYUv2, and has comparable performance on the other tasks.}
\end{table}
\subsection{Multi-task learning}
We test GREAT for multi-task learning on 2 datasets: (a) CIFAR-10 with input a noisy gray-scale image and with tasks (i) classification, (ii) colorization, (iii) edge detection and (iv) denoised reconstruction; (b) NYUv2 dataset wherein the tasks are (i) depth estimation, (ii) surface-normal estimation, and (iii) key-point estimation. The input and output resolutions for the CIFAR-10 dataset are 32x32, and the input resolution for NYUv2 is 320x320 and the output resolution is 80x80 as set in \cite{gradnorm}. We compare out method against the baseline of equal weights, GradNorm \cite{gradnorm}, and uncertainty based weighting \cite{kendallmulti}. For all methods we use the same architecture: a ResNet-53 with dilated convolution backbone and task-specific decoders. We tested GradNorm for different $\alpha$ values and set it equal to 0.6 for CIFAR-10, and 1.5 for NYUv2 as set in the original paper. Full details about the dataset creation, task losses, main model and classifier architecture are in the supplement. Table~\ref{multitask} lists the results. We see that 
GREAT performs better or on par with GradNorm, despite having no tunable hyperparameters. This indicates that the extra parameters in the GALs are sufficient to absorb dataset-specific information without requiring hand-tuning. 
On CIFAR-10, we see that GREAT performs best on edge detection and denoised auto-encoding, and is close to the best value for colorization. The high classification error for the uncertainty-based method and high RMSE values of the baseline on the other three tasks indicates that classification is antagonistic to the other three tasks. However, both GradNorm and GREAT are able to correctly balance the gradient flowing from classification with the other tasks. On the NYUv2 dataset we see that GREAT performs best on depth and normal estimation, and is within $\approx 0.005$ RSME on keypoint detection. Overall, we see that GREAT performs better than all other methods on four of the seven tasks, and is close to the best values in all cases. 

\section{Conclusion and future work}
We have introduced gradient adversarial training and demonstrated its applicability in diverse scenarios: from defense against adversarial examples to knowledge distillation to multi-task learning. We show that adaptations of GREAT offer (a) strong defense to both targeted and non-targeted adversarial examples, (b) can easily distill knowledge from different teacher networks without heavy parameter tuning, and (c) aid multi-task learning by tuning a gradient alignment layer. There are several directions of future work in the proposed domains. We wish to investigate others forms of loss functions beyond GREACE that are symbiotic with GREAT, explore progressive training of student networks using ideas from Progressive-GAN \cite{progressivegan} to better learn from the teacher, and absorb the explicit parameters in the GALs directly into the optimizer as done with the mean and variance estimates for each weight parameter in ADAM~\cite{adam}. The general approach underlying GREAT of passing an adversarial gradient signal to a network is broadly applicable to domains beyond the ones discussed here such as to the discriminator in domain adversarial training~\cite{DANN} and GANs~\cite{GAN}. We can also replace direct gradient tensor evaluation with synthetic gradients~\cite{syntheticgrad} for efficiency. In the future we will explore these exciting avenues. Holistically, we believe that understanding gradient distributions will help uncover the underlying mechanisms that govern the successful training of deep architectures using backpropagation, and  gradient adversarial training is a step towards this direction.



\bibliographystyle{plain}
\bibliography{sample.bib}

\newpage
\subsubsection*{\Large{Supplementary Material}}

\textbf{GREAT for adversarial defense}
 
 We use the publicly available CIFAR-10 and SVHN datasets for our experiments. Both datasets have color images of size 32x32x3 and 10 classes. CIFAR-10 has 50000 training examples and 10000 test examples. SVHN has 73257 digits for training and 26032 digits for testing. We perform data augmentation by 4-padding the image during training and randomly cropping a 32x32 image. As mentioned in the main manuscript we use a ResNet-18 architecture for the main network on both datasets. We use a ResNet-18 for the auxiliary classifier as well, with the ReLU activations replaced with leaky-ReLU. The $\alpha$ parameter in leaky-ReLU is set to 0.2.  We use ADAM with initial learning rate of 0.001 to train both the main and auxiliary network, and $lr$, the learning rate multiplier is calculated as $(1-{e}/{e_{max}})^{0.9}$ where $e,e_{max}$ are the current epoch and total epochs, respectively. $\alpha$ and $\beta$ follow the rate policy of $\alpha_{max}(1-lr)$ and $\beta_{max}(1-lr)$, i.e., increase gradually with epochs. 
 
 The adversarially trained network is fed clean samples and adversarial samples with equal probability. The total loss used for training the network is $0.5*(L_{clean}+L_{adv})$, where $L_{clean}$ and $L_{adv}$ is the cross-entropy loss on the true labels. Adversarial examples are generated using a single-step FGSM and $\epsilon=0.2$ for both datasets. We train all networks for 100 epochs on both datasets.

  \textbf{GREAT for knowledge distillation}

As described in the main manuscript, we evaluate knowledge distillation for four scenarios. We describe the architecture for student, teacher and auxiliary network for each scenario along with other hyper-parameters used during training.
\begin{itemize}
\item CNN-5+ResNet-18: We use a 5-layer CNN model with 3 convolution layers of kernel size 3x3 and padding 1. The second and third convolution layers are of stride 2. The final two layers are linear layers. The number of channels in the first layer are 32 and it increases by a factor of 2 in every succeeding convolutional layer.  The penultimate  layer has 128 channels. All layers are succeeded by batch-normalization and ReLU activation, except the last linear layer. The teacher is a pre-trained ResNet-18 architecture on CIFAR-10 obtained from 
\footnote{\url{https://stanford.app.box.com/s/5lwrieh9g1upju0iz9ru93m9d7uo3sox}}. The binary classifier is of the same architecture as the student network, except the ReLU activations are replaced with leaky-ReLU with $\alpha=0.2$. We perform data augmentation during training in the form of random horizontal flips and random crops from a 4-padded image. We use ADAM optimizer with initial learning rate of 0.001 decreased using the multiplier $(1-{e}/{e_{max}})^{0.9}$ as described for adversarial defense. We train the network for 50 epochs on both the dense as well as sparse sample regime. 
\item ResNet-18+ResNext-29-8: We use a a standard ResNet-18 as the student model and a pre-trained ResNext-29-8 architecture on CIFAR-10 obtained from the same link as above. The binary classifier is a ResNet-11 architecture comprised of 8 ResNet module layers, 1 convolutional layer, 1 average pooling layer and final linear layer for classification. The ReLU activations in the ResNet blocks are replaced with leaky-ReLU ($\alpha=0.2$ ). We perform data augmentation during training in the form of random horizontal flips and random crops from a 4-padded image. We use ADAM optimizer with initial learning rate of 0.001 decreased using the multiplier $(1-{e}/{e_{max}})^{0.9}$ as described for adversarial defense. We train the network for 100 epochs on both the dense as well as sparse sample regime. 
\item ResNet-18+ResNet-50 : We use a a ResNet-18 and ResNet-152 suitable for CIFAR-10 dataset as the student-teacher combination. The first convolutional layer is of stride 2 to compensate for 64x64 input size. The binary classifier is a ResNet-18 like ResNet-11 architecture comprised of 4 ResNet module layers each with 2 convolutional layers, 1 simple convolutional layer, 1 average pooling layer and final linear layer for classification. The first layer in the ResNet-11 architecture is of stride 2 to compensate for 64x64 input size. The ReLU activations in the ResNet blocks are replaced with leaky-ReLU($\alpha=0.2$ ). The images are randomly scaled and resized to have minimum dimension of 70. Then a random crop of size 64x64 is performed on the augmented images. We use SGD optimizer with momentum 0.9, weight decay  5e-4, and initial learning rate of 0.01 decreased using the multiplier $(1-{e}/{e_{max}})^{0.9}$. We train the teacher network for 100 epochs. We then train the student model for 50 epochs on both the dense as well as sparse sample regime. 
\item ResNet-18+ResNet-152 : We use a a ResNet-18 and ResNet-152 suitable for ImageNet as the student-teacher combination.  The binary classifier is a ResNet-18 like ResNet-11 architecture comprised of 4 ResNet module layers each with 2 convolutional layers, 1 simple convolutional layer, 1 average pooling layer and final linear layer for classification. The ReLU activations in the ResNet blocks are replaced with leaky-ReLU ($\alpha=0.2$ ). The images are randomly scaled and resized to have minimum dimension of 256. Then a random crop of size 224x224 is performed on the augmented images. We use SGD optimizer with momentum 0.9, weight decay  5e-4, and initial learning rate of 0.01 decreased it using the multiplier $(1-{e}/{e_{max}})^{0.9}$. We train the teacher network for 50 epochs. We then train the student model for 30 epochs on both the dense as well as sparse sample regime. 
\end{itemize}

We set $\alpha=0.1$ and temperature value equal to 20 for distillation in the dense regime. We set $\alpha=0.99$ and temperature value equal to 20 for distillation in the sparse regime. These values are optimal for CNN-5+ResNet-18 combination as determined via grid search. The same values were used for experiments on mini-ImageNet. The optimal values for the sparse and dense regime for the ResNet-18+ResNext-29-8 combination were determined to be $\alpha=0.95$ and temperature equal to 6. We also investigated Sobolev training of neural networks which directly matches the gradient distribution on the CNN-5+ResNet-18 combination. We determined optimal alpha parameters, and evaluated the dense and sparse regime accuracy to be 84.98 and 66.50\% percent respectively, as opposed to GREAT with 85.72\% and 66.55\% respectively.

  \textbf{GREAT for multitask learning}
  
 We perform multi-task learning on two datasets: (a) CIFAR-10 and (b) NYUv2. For CIFAR-10 we convert the color images into grayscale. We then add  salt and pepper noise with equal probability to 4\% of the pixels. We further add speckle noise with gaussian probability of 0.1. We use a Canny edge detector with $\sigma=1$ on the colored images and set it to be the ground truth edges.  The shared encoder in the CIFAR-10 multi-task learning architecture is a standard ResNet-18 minus the pooling and linear layers. These layers are instead used in the classification decoder. The decoder for edge detection, colorization and denoising consist of 3 upsampling layers of scale 2 and 3 ResNet blocks. 
Classification task was trained with cross-entropy loss and the other three were trained using MSE loss. 
We use ADAM optimizer with initial learning rate of 0.001 and decreased it using the multiplier $(1-{e}/{e_{max}})^{0.9}$. The classifier is a ResNet-11 with 256 input channels, decreased by a factor of 2 over 4 ResNet modules, followed by a pooling and a linear layer for 4-way classification.  The ReLU activations in the ResNet blocks are replaced with leaky-ReLU with $\alpha=0.2$. The weight tensors (GradNorm, Uncertainty), task classifier and GALs were trained with ADAM and the same learning rate policy as the main network. Note we normalize the weights in the uncertainty based method to sum to 1 for a fair comparison to GradNorm. We train the networks for 50 epochs.

The multitask dataset for NYUv2 is created from the NYUv2 depth images. We augment the standard NYUv2 depth dataset  with additional frames from each video, resulting in $\approx$45,000
images complete with pixel-wise depth, surface normals, and room keypoint labels. Keypoint
labels are obtained through professional human labeling services, while surface normals are generated algorithmically. The full dataset is then split by scene and we get 26328 training images and 18582 test images. All inputs are downsampled to 320 x 320 pixels and outputs to 80 x 80 pixels. We use these resolutions following GradNorm. The shared encoder is a dilated residual network with 43 layers, of which 2 intermediate layers have a dilation rate of 2 and 1 has a dilation rate of 4. The input is downsampled 8 times over the layers. The number of output channels after the encoder is 512. The decoder for the 3 tasks comprises of a single upsampling of stride 2 and 3 convolution layers. Depth estimate has a single channel, normal estimation has 3 channel, while keypoint estimation has 48 channels. As in GradNorm, we generate Gaussian heatmaps for each of 48 room keypoint types and predict these heatmaps with a pixel-wise squared loss. We use one minus the absolute cosine similarity as the normal estimation loss and MSE as the depth estimation loss. The classifier is a ResNet-11 with 512 input channels, decreased by a factor of 2 over 4 ResNet modules, followed by a pooling and a linear layer for 3-way classification.  The ReLU activations in the ResNet blocks are replaced with leaky-ReLU with $\alpha=0.2$.  We use ADAM optimizer with initial learning rate of 0.001 and decreased it using the multiplier $(1-{e}/{e_{max}})^{0.9}$ for training all modules and networks. We train the network for 30 epochs.

\end{document}